\documentclass[runningheads]{llncs}
\usepackage[T1]{fontenc}
\usepackage{graphicx}
\usepackage{amsmath}
\usepackage{amssymb}
\usepackage{epstopdf}
\usepackage{color}
\usepackage{multirow}
\usepackage{marvosym}
\newcommand\UrlFont{\color{blue}\rmfamily}

\begin{document}
\title{XMorpher: Full Transformer for Deformable Medical Image Registration via Cross Attention}
\titlerunning{Full Transformer for Deformable Image Registration}
%
\author{Jiacheng Shi\inst{1} \and Yuting He\inst{1} \and
Youyong Kong\inst{1,2,3} \and Jean-Louis Coatrieux\inst{2,3} \and Huazhong Shu\inst{1,2,3} Guanyu Yang\inst{1,2,3}\textsuperscript{(\Letter)} Shuo Li\inst{4}}
\authorrunning{J. Shi et al.}
%
\institute{LIST, Key Laboratory of Computer Network and Information Integration (Southeast University), Ministry of Education, Nanjing, China \and Jiangsu Provincial Joint International Research Laboratory of Medical Information Processing \and Centre de Recherche en Information Biomedicale Sino-Franc, ais (CRIBs) \and Dept. of Medical Biophysics, University of Western Ontario, London, ON, Canada\\
\email{yang.list@seu.edu.cn}}
\maketitle              
\begin{abstract}
An effective backbone network is important to deep learning-based Deformable Medical Image Registration (DMIR), because it extracts and matches the features between two images to discover the mutual correspondence for fine registration. However, the existing deep networks focus on single image situation and are limited in registration task which is performed on paired images. Therefore, we advance a novel backbone network, XMorpher, for the effective corresponding feature representation in DMIR. \textbf{1)} It proposes  a novel full  transformer architecture including dual parallel feature extraction networks which exchange information through cross attention, thus discovering multi-level semantic correspondence while extracting respective features gradually for final effective registration. \textbf{2)} It advances the Cross Attention Transformer (CAT) blocks to establish the attention mechanism between images which is able to find the correspondence automatically and prompts the features to fuse efficiently in the network. \textbf{3)} It constrains the attention computation between base windows and searching windows with different sizes, and thus focuses on the local transformation of deformable registration and enhances the computing efficiency at the same time. Without any bells and whistles, our XMorpher gives Voxelmorph $2.8\%$ improvement on DSC , demonstrating its effective representation of the features from the paired images in DMIR. We believe that our XMorpher has great application potential in more paired medical images. Our XMorpher is open on \UrlFont{https://github.com/Solemoon/XMorpher}
\end{abstract}
\section{Introduction}
A powerful backbone network is important to deep learning (DL)-based Deformable Medical Image Registration (DMIR)\cite{balakrishnan2018unsupervised,de2019deep,sotiras2013deformable}. The backbone network is able to extract the features of moving and fixed images in DMIR and then 
\begin{figure}
\includegraphics[width=\textwidth]{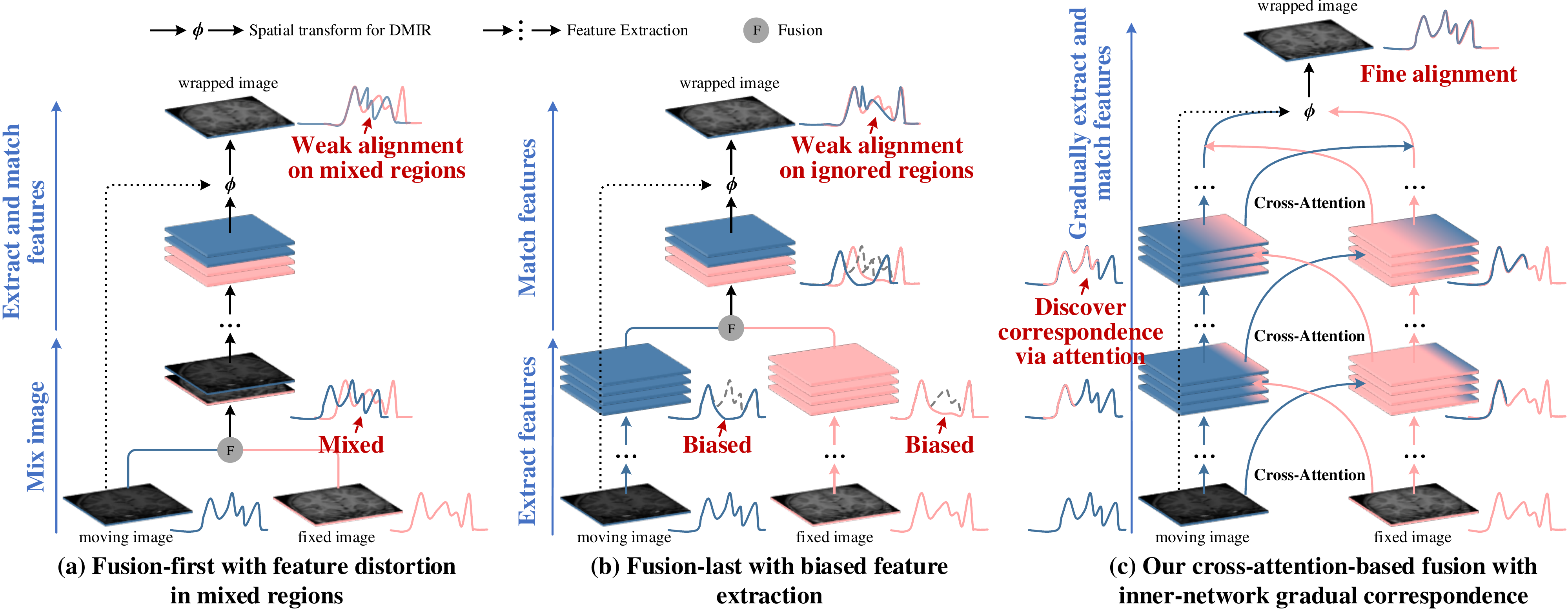}
\caption{The limitations of existing methods and our superiority. (a) Fusion-first with feature distortion in mixed regions making inaccurate representation for corresponding features. (b) Fusion-last with biased feature extraction losing correspondence of ignored features. (c) Our cross-attention-based fusion with inner-network gradual correspondence for fine alignment of different level features.} \label{fig1}
\end{figure}
match the features to obtain correspondences from moving images to fixed images, thus contributing to a fine registration with these effective correspondences. The moving images are transformed into the same coordinate system as the fixed images after registration, which makes it convenient and efficient to compare different images for doctors,  thus greatly promoting the efficiency of diagnosis and reducing the cost of disease.

Although existing deep networks\cite{ronneberger2015u,tajbakhsh2016convolutional} have strong performance in single image feature representation, these Single Image Networks (SINs) are still limited in feature extraction and match of a pair of images in DMIR: \textbf{1)} \emph{Fusion-first with feature distortion in mixed regions.} As shown in Fig.\ref{fig1}(a), some DMIR methods\cite{balakrishnan2018unsupervised,balakrishnan2019Voxelmorph,vos2017end} fuse the moving and fixed images to simulate the single-image input condition, and the fusion is sent to a SIN for moving-fixed features. But these methods mix the feature extraction and feature matching processes together, leading to the feature distortion and weak alignment in the mixed regions, thus making the networks unable to identify one-to-one correspondences between the image pairs. The inefficient capacity of feature representation finally contributes to the absence of critical structures and poor registering details. \textbf{2)} \emph{Fusion-last with biased feature extraction.} As shown in Fig.\ref{fig1}(b), these networks\cite{klein2009elastix} send moving and fixed images to dual SINs respectively and fuse features from different networks at the end. But these networks separate the feature extraction and feature matching processes absolutely, leading to final match of two biased features from different SINs, thus making the networks ignore the different levels (such as multiscale) of the features in some regions. The unicity of feature representation limits the correspondence of different information between images and leads to poor registration finally.

The attention mechanism of transformer\cite{han2022survey} provides the potential application in registration because of its outstanding capacity of catching relevance in images, but existing researches of transformer\cite{chen2021transunet,khan2021transformers,liu2021swin,zheng2021rethinking} only focus on single image scenarios and lack related design for moving-fixed correspondence between two images in DMIR. These transformers utilize the self-attention function to obtain an output with the weight information which points out the areas needing to be focused in one image. This mechanism digs out the relationships of internal basic elements and extracts the most task-related features, and thus transformer has a good performance in single-image tasks and is potential in DMIR. \textbf{However}, current transformers\cite{chen2021transmorph,10.1007/978-3-030-87202-1_13} for DMIR still take the same attention mechanism as single-image tasks which focus on the relevance in one image but ignore the correspondences between image pairs. The competence of capturing correspondences between moving and fixed images restrains transformers to find effective registering features for fine registration.

In this paper, we proposed a novel transformer, X-shape Morpher (XMorpher) for dual images input in DMIR, it advances Cross Attention to the transformer architecture for efficient and multi-level semantic feature fusion, thus effectively improving the performance of registration. In short, the contributions of our work are summarized as follows: 
\textbf{1)} We proposed a novel full  transformer backbone network for DMIR. As shown in Fig.\ref{fig1}(c), it includes dual parallel feature extraction sub-networks whose respective features are fused and matched in the form of cross attention. Through the progressive and commutative network, the features from different images are fused and matched gradually through the cross-attention-based fusion modules, thus achieving effective feature representation of moving-fixed correspondences and gaining fine-grained multi-level semantic information for fine registration.
\textbf{2)} We present a new attention mechanism, Cross Attention Transformer (CAT) block, for sufficient communication between a pair of features from moving and fixed images. CAT block utilizes the attention mechanism to compute the mutual relevance, thus learning the correspondences between two images and promoting the features to match automatically in the network.
\textbf{3)} We constrain the feature matching process in windows based on the local transformation of DMIR, which narrows the searching range between  moving and fixed images, thus increase the computation efficiency and reduce the interference from similar structure if matching in a large space. This window-based feature communication greatly improve the accuracy and efficiency of registration.
\section{Methodology}
\begin{figure}[!htb]
\includegraphics[width=\textwidth]{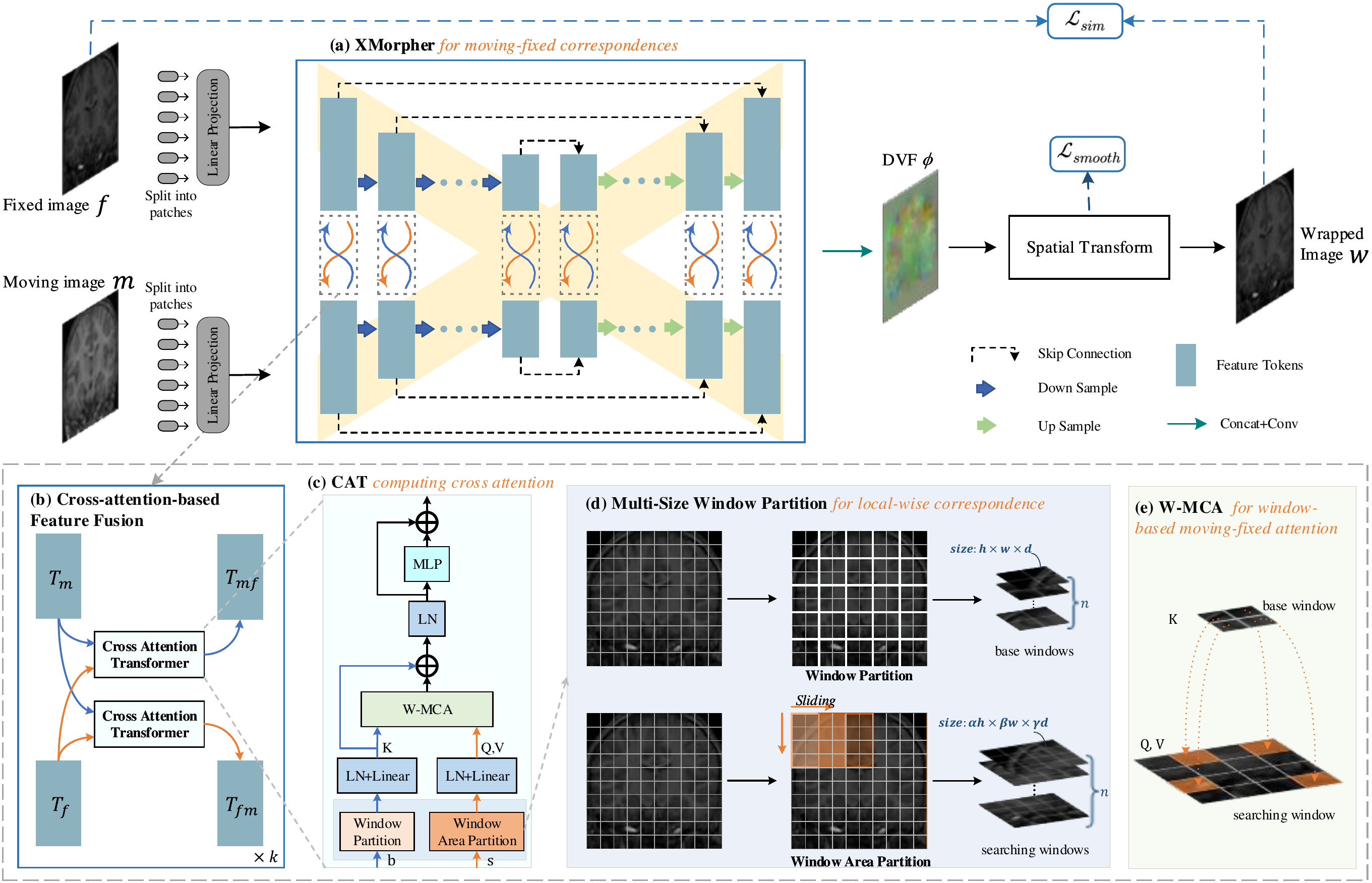}
\caption{Overall architecture of our XMorpher. (a) It includes dual U-shape networks which exchange features through our cross-attention-based feature fusion. (b) The feature fusion module is composed of two CAT blocks sharing parameters for mutual correspondences. (c) The detailed construction of our CAT block which fuses two input features into one with attention information. (d) Our two different methods of window partition to generate base and searching windows. (e) We compute the cross attention between base and searching windows through W-MCA.} \label{fig2}
\end{figure}

As is shown in Fig.\ref{fig2}, XMorpher  is based on DL-based registration networks\cite{balakrishnan2019Voxelmorph,vos2017end}, and it is used to extract and match moving and fixed features in registration for effective representation of input image pairs. The final representation generates the DVF $ \phi $ and obtain the fine wrapped image $ w $ through spatial transform\cite{he2020deep}. Our XMorpher includes: \textbf{1)} A X-shape transformer architecture with dual parallel U-shape feature extraction sub-networks which keeps exchanging information through cross-attention-based feature fusion modules. \textbf{2)} CAT blocks for feature fusion between two features tokens from different sub-networks. \textbf{3)} Multi-size window partition in CAT for  precise local-wise correspondences.
\\\\
\textbf{2.1 XMorpher for efficient and multi-level semantic feature representation in registration} \\
\textbf{I. X-shape architecture with parallel communicating feature extraction networks.}
As shown in Fig.\ref{fig2}(a), we utilize dual U-shape networks to extract the features of moving and fixed images respectively and the two networks communicate through feature fusion modules,  thus forming a X-shape network, so we name it as XMorpher. The two parallel networks follow the structure of Unet\cite{ronneberger2015u} with encoding and decoding parts, but we replace the convolutions with our CAT blocks which play important role in the attention-wise feature fusion modules (Fig.\ref{fig2}(b)) between the two networks. Through the parallel communicating networks, our XMorpher exchange cross-image information vertically and keeps refining features horizontally. So the final output features have strong ability of representing the correspondences between moving and fixed images. \\
\textbf{II. Cross-attention-based feature fusion module.}
As shown in Fig.\ref{fig2}(b), the corresponding features $T_{m}$ and $T_{f}$ coming from the parallel sub-networks obtain their mutual attention through two CAT blocks by exchanging the order of inputs. Then the two outputs with the other's attention return to the original pipelines and prepare for next deeper communication. There are $k$ times of communication in total in a feature fusion module for sufficient mutual information. Through the attention-wise feature fusion modules between two networks, features from different networks with different semantic information communicate frequently, thus our XMorpher keeping learning multi-level semantic features for final fine registration.\\\\\\
\textbf{2.2 Cross Attention Transformer block for corresponding attention}\\
CAT block aims to compute new feature tokens with corresponding relevance from input feature $b$ to feature $s$ through the attention mechanism. As shown in Fig.\ref{fig2}(c), $b$ and $s$ are respectively partitioned in different ways (described in section 2.3) into two sets of windows, base window set  $ S_{ba} $ and searching window set $ S_{se} $, for next window-based attention calculation.  $ S_{ba} $ and $ S_{se} $ have the same size $ n $ and different window sizes. Each base window in $ S_{ba} $ is projected to the query set $query$ and each searching window is projected to knowledge set $key$ and $value$ through linear layer. Then our Window-based Muti-head Cross Attention (W-MCA) compute the cross attention between two windows and the attention is added to the base window so that each base window gets the corresponding weighed information from searching window. Finally, the new output set is sent to a 2-layer MLP\cite{vaswani2017attention} with GELU non-linearity to enhance learning ability. A LayerNorm (LN) layer is applied before each W-MCA and each MLP module to guarantee validity of each layer.\\\\
\textbf{2.3 Multi-size window partitions for local-wise correspondence}\\
\textbf{I. Window Partition (WP) and Window Area Partition (WAP). }
Since the deformable image registration focuses on local displacement of voxel and there is no large span correspondence between moving and fixed images, we proposed the window-based cross attention mechanism which utilizes multi-size window partitions to limit attention computation in windows. Multi-size window partitions include two different methods, WP and WAP, to divide the input feature tokens $b$ and $s$ into windows of different sizes. As shown in Fig.\ref{fig2}(d), WP partition feature tokens directly into base window set $ S_{ba} $ with size of $ n\times h\times w\times d $ and WAP enlarges the window size with the magnifications $\alpha$, $\beta$ and $\gamma$. So the base and searing window size are calculated as:
\begin{equation}
\begin{split}
&h_{ba},w_{ba},d_{ba}= h,w,d \\
&h_{se},w_{se},d_{se}=\alpha \cdot h,\beta \cdot w,\gamma \cdot d
\end{split}
\end{equation}
where $ h_{ba}, w_{ba},d_{ba} $ are the size of base windows and $ h_{se}, w_{se},d_{se} $ are the size of searching windows. To obtain the same amount of two window sets, WAP takes advantage of a sliding window and the stride is set as the base window size, and thus $ S_{se} $ has size of $ n\times \alpha \cdot h \times \beta \cdot w \times\gamma \cdot d $. Through the corresponding windows with different sizes, CAT blocks compute the cross attention between two feature tokens efficiently and avoid large-span searches for precise correspondence.\\
\textbf{II. Window-based Muti-head Cross Attention (W-MCA).}
We proposed W-MCA to compute the cross attention between acquired base windows and searching windows to find the mutual correspondences.  W-MCA is a function mapping a query and a set of key-value pairs to an output, where the query comes from the base windows, keys and values come from the searching windows. The output is computed as a weighted sum of the values, where the weight assigned to each value is computed by a compatibility function of the query with the corresponding key. W-MCA adopts multi-head attention\cite{vaswani2017attention} for ample representation subspaces. W-MCA computes the dot products of query and keys and applies a softmax function to obtain the weights on the values. So our cross attention is computed as:
\begin{equation}
W-MCA\left( Q_{ba},K_{se},V_{se}\right)  =softmax\left( \frac{Q_{ba}K^{T}_{se}}{\sqrt{d} } \right)  V_{se}
\end{equation}
where $ Q_{ba}, K_{se}, V_{se} $ are the $ query,key$ and $ value $ matrices. $ Q_{ba}\in \mathbb{R}^{n\times s\times c} $ is the linear projection of $ S_{ba} $ and $ K_{se},V_{se}\in \mathbb{R}^{n\times \mu\cdot s\times c} $ are linear projections of $ S_{se} $, $ s=h\times w\times d $ and $ \mu =\alpha \cdot \beta \cdot \gamma $, and $ c $ is the dimension of each feature token. 
\section{Experiment}
\textbf{3.1 Experiment protocol}\\
We perform effective experiments to evaluate our XMorpher's superiority both in unsupervised and semi-supervised strategies. \textbf{1)} \emph{Dataset.} We validate the performance of our XMorpher on  the whole heart registration tasks on the CT dataset from MM-WHS 2017 Challenge\cite{zhuang2016multi} which has 20 labeled images and 40 unlabeled images and ASOCA\cite{ramtin_gharleghi_2020_3819799} which has 60 unlabeled images. We use all the unlabeled images (100 images) and 5 labeled images to compose 500 labeled-unlabeled image pairs and 9900 unlabeled-unlabeled image pairs as the training set for unsupervised and semi-supervised experiments. The remaining 15 of the labeled images compose 210 image pairs as testing set. All images were preprocessed to the same spatial coordinates through aﬃne transformation\cite{tustison2014large}. \textbf{2)} \emph{Implementation.} We apply our XMorpher as backbone in two registration frameworks: unsupervised Voxelmorph\cite{balakrishnan2018unsupervised} (VM-XMorpher) and semi-supervised PC-Reg\cite{he2021few} (PC-XMorpher). Furthermore, the hyperparameters of XMorpher is set as $ h=w=d=2 $ and $ \alpha =\beta =\gamma =3 $. The proposed framework is  implemented by PyTorch on NVIDIA GeForce RTX 3090 GPUs with 24 GB memory. \textbf{3)} \emph{Comparison settings.} We set up a set of comparative experiments to prove the advancement of our XMorpher. Controlled experiments include BSpline\cite{rueckert1999nonrigid}, Voxelmorph\cite{balakrishnan2018unsupervised}, PC-Reg\cite{he2021few}, Transmorph\cite{chen2021transmorph}. \textbf{4)} \emph{Evaluation metrics.} We use the mean dice similarity coeﬃcients (DSC) of all labels to evaluate the performance of registration and the Jacobian matrix ($ \left| J\left( \psi \right)  \right|  \leqslant 0$ (\%)) to evaluate the rationality of registration fields. 
\\
\begin {table}
\begin{center}
\caption{The proposed XMorphers achieve the state-of-the-art performance on DSC both under unsupervised and semi-supervised strategies, as well as have top-ranked performance on Jacobian matrix ($|J_{\phi}|\leq0$ (\%)).}\label{tab1}
\begin{tabular}{lcccc}
\hline
Method                                          &Un-/Semi-              &Backbone           &DSC                    &$|J_{\phi}|\leq0$ (\%) \\
\hline
Affine initialization                           &-                      &-                  &$69.2\pm7.2$           &-\\
\hline
BSpline \cite{rueckert1999nonrigid}             &\multirow{5}{*}{Unsup} &-                  &$80.9\pm7.6$           &$5.25\pm3.27$\\
Voxelmorph \cite{balakrishnan2018unsupervised}  &                       &CNN                &$80.2\pm 5.5$          &$4.02\pm0.82$ \\
Transmorph \cite{chen2021transmorph}            &                       &CNN+Transformer    &$81.1\pm 5.2$          &$3.46\pm0.75$ \\
\textbf{Our no-cross XMorpher}                  &                       &Full Transformer   &$81.5\pm5.4$           &\textbf{0.94$\pm$0.26}\\
\textbf{Our VM-XMorpher}                     &                       &Full Transformer   &\textbf{83.0$\pm$4.7}  &$3.15\pm0.79$\\
\hline
PC-Reg\cite{he2021few}                          &\multirow{2}{*}{Semi}  &CNN                &$86.0\pm2.5$           &$0.36\pm0.20 $ \\
\textbf{Our PC-XMorpher}                      &                       &Full Transformer   &\textbf{86.9$\pm$2.4}  &\textbf{0.32$\pm$0.18}\\
\hline
\end{tabular}
\end{center}
\end{table}
\begin{figure}[!htb]
\includegraphics[width=\textwidth]{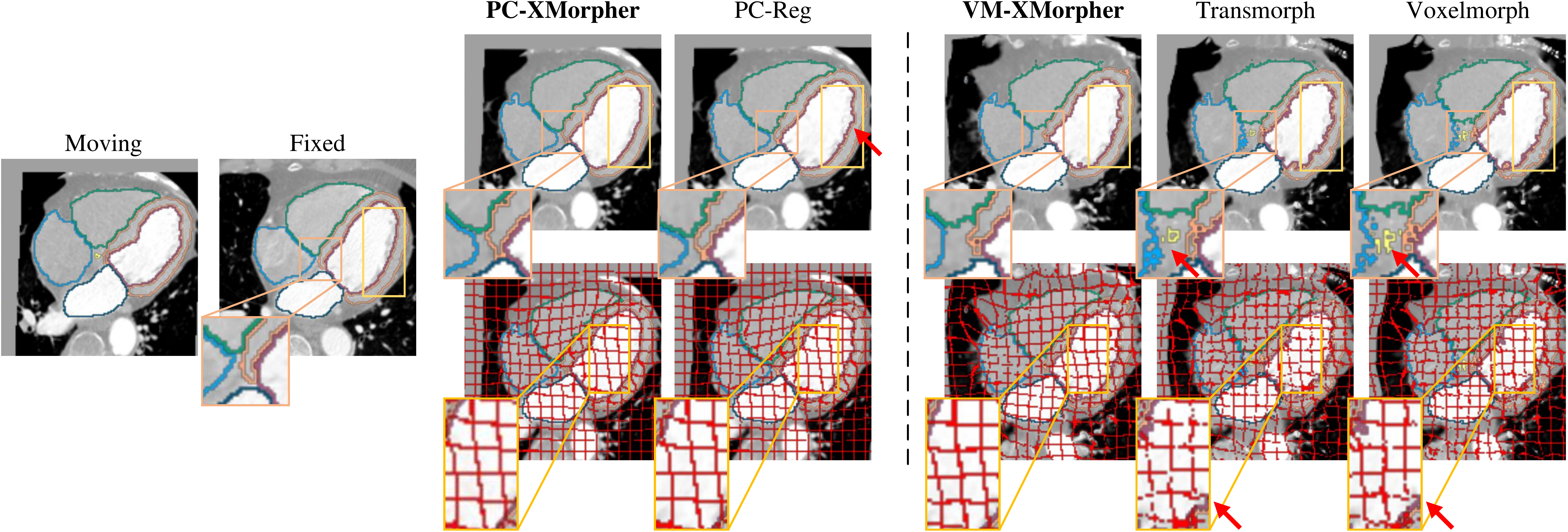}
\caption{The visual results of Voxelmorph, Transmorph, PC-Reg and ours. XMorphers have obvious visual advantages both in unsupervised and semi-supervised strategies.} \label{fig3}
\end{figure}
\\
\textbf{3.2 Results and analysis} 
\\
\textbf{Quantitative comparison.}
Our VM-XMorpher achieves the highest DSC score of $83.0\%$ and has the top-ranked performance on Jacobian matrix at the same time (tab.\ref{tab1}), which shows that it not only has a strong registration effect, but also has a strong maintenance of the image structure. VM-XMorpher  is $2.8\%$ higher on DSC and  $0.87\%$ lower on Jacobian matrix than Voxelmorph\cite{balakrishnan2018unsupervised}, illustrating that XMorpher represents the correspondence effectively and thus results in refined details and strong topology preservation compared with fusion-first networks. VM-XMorpher still has an obvious improvement in contrast to Transmorph\cite{chen2021transmorph} which combines the CNN and transformer, evaluating that cross attention mechanism has an superiority over the existing transformers and full transformer is good at focus on the cross-image correspondence. Furthermore, PC-Reg has achieved considerable performance on registration while our PC-XMorpher is still $0.9\%$ higher on DSC and has better performance on Jacobian matrix. These results prove the superiority of our XMorpher under different training strategies.
\\
\textbf{Visual superiority.}
\begin{figure}
\includegraphics[width=\textwidth]{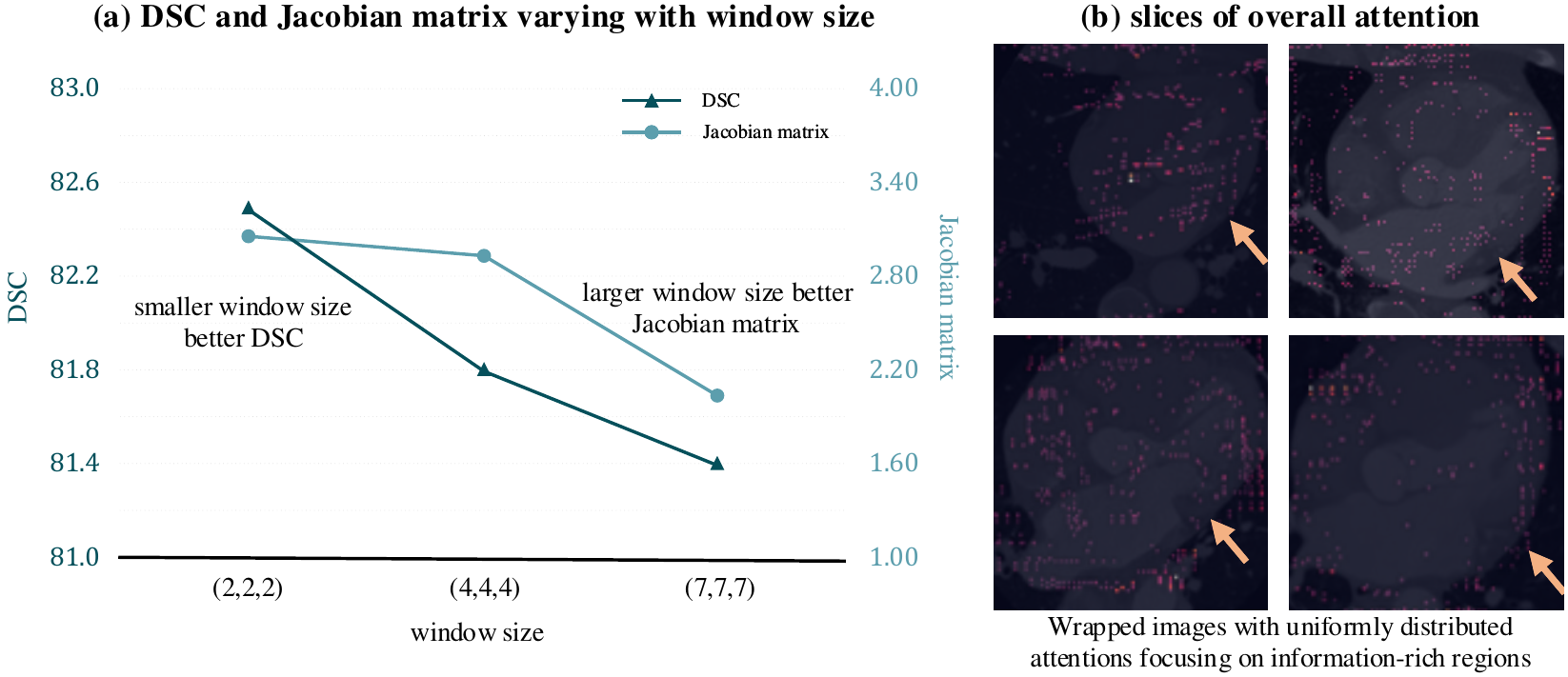}
\caption{(a) The DSC and Jacobian matrix of XMorpher have inverse variation with the window size. (b)The uniformly distributed window attentions indicate that XMorpher effectively find the keys in every windows and the keys correspond to the heart region.} \label{fig4}
\end{figure}
Fig.\ref{fig3} shows one case of the visual results of our XMorpher and other compared experiments in different training strategies. Our PC-XMorpher has better details compared with PC-reg such as the more smooth boundaries. Furthermore, our XMorpher has obvious visual superiority under unsupervised experiments.VM-XMorpher still has much better ability of boundary recognition than other methods and stronger resolution of the neighbouring regions while other models (Voxelmorph, Transmorph) have the obvious mixture of two borders. Furthermore, the deformation grids of VM-XMorpher is also smooth compared with other methods which have many cracks in the grids.
\\
\textbf{Ablation study.}
Through the ablation experiments, we demonstrated the effect of CAT blocks and the influence of the window size. \textbf{a)} Ablation for CAT blocks. We designed no-cross XMorpher whose input is the concatenation of moving and fixed images. Its transformer blocks computes the internal attention of the concatenation and the other components are same as XMorpher. Our VM-XMorpher has 1.5 percent increase on DSC demonstrating that the effective correspondence representation enhances the interaction between the two images in DMIR but brings the risk of less smooth deformation at the same time. But our VM-XMorpher still has better performance on Jacobian matrix than other models. \textbf{b)} Ablation for window size. Fig.\ref{fig4}(a) demonstrates that the larger windows has better performance on the Jacobian matrix because the more structure information of the images will be extracted in a wider horizon, but the smaller windows result in a better DSC owing to finer searching space.
\\
\textbf{Analysis of visual cross attention.}
The uniformly distributed visual attention (Fig.\ref{fig4}(b)) illustrates that the window-based attention mechanism disperses the correspondence into windows and takes advantage of these uniform mappings of small regions for a fine registration. Furthermore, the high-attention regions with ample information correspond to the heart region, demonstrating the powerful image comprehensive ability of our XMorpher.
\section{Conclusion}
We propose a full transformer network XMorpher which effectively represents multi-level correspondence between moving and fixed images. Furthermore, we use window-based CAT blocks to extract and match the features in the network efficiently to obtain final cross-image correspondences for fine registration. The compared experiments have prove that XMorpher has outstanding performance in DMIR with different training strategies and we believe that it has great application prospect in diagnosis and treatment. 
\subsubsection{Acknowledgment}
This work was supported in part by the National Key Research and Development Program of China (No.2021ZD0113202), in part by the National Natural Science Foundation under grants (61828101), CAAI-Huawei MindSpore Open Fund, CANN(Compute Architecture for Neural Networks), Ascend AI Processor, and Big Data Computing Center of Southeast University.
%
%
\bibliographystyle{splncs04}
\bibliography{references}

\end{document}